%% file: naaclhlt2019.tex
\documentclass[11pt,a4paper]{article}
\usepackage[hyperref]{naaclhlt2019}
\usepackage{times}
\usepackage{latexsym}

\usepackage{url}
\usepackage{enumitem}
\usepackage{pifont}
\newcommand{\cmark}{\ding{51}}%
\newcommand{\xmark}{\ding{55}}%
\usepackage{textcomp}

\usepackage{graphicx}
\usepackage{subfigure}
\usepackage[font={footnotesize}]{caption}

\newcommand\csqa{\textsc{CommonsenseQA}}
\newcommand\dataset{\textsc{CommonsenseQA}}
\newcommand\conceptnet{\textsc{ConceptNet}}
\newcommand\nlex[1]{\emph{``{#1}"}}

\usepackage{amsmath,amssymb}

\usepackage{varwidth}

\aclfinalcopy 

\newcommand\ignore[1]{}

\input{std-macros}

\title{\csqa{}: A Question Answering Challenge Targeting Commonsense Knowledge}

\author{Alon Talmor\thanks{\;\;\; The authors contributed equally}$^{*,1,2}$ ~~~~~ Jonathan Herzig$^{*,1}$ ~~~~~
Nicholas Lourie$^2$ ~~~~~
Jonathan Berant$^{1,2}$ \\
\mbox{}\\
$^1$School of Computer Science, Tel-Aviv University \\
$^2$Allen Institute for Artificial Intelligence \\
\small{\texttt{\{alontalmor@mail,jonathan.herzig@cs,joberant@cs\}.tau.ac.il},~\texttt{nicholasl@allenai.org}}}
\date{}

\begin{document}
\maketitle
\begin{abstract}
\input abstract
\end{abstract}

\input introduction
\input related_work
\input dataset_gen

\input dataset_analysis

\input baselines

\input experiments

\input conclusion

\section*{Acknowledgments}
We thank the anonymous reviewers for their constructive feedback. This work was completed in partial fulfillment for the PhD degree of Jonathan Herzig, which was also supported by a Google PhD fellowship. This research was partially supported by The Israel Science Foundation grant 942/16, The Blavatnik Computer Science Research Fund and The Yandex Initiative for Machine Learning.

\bibliography{all}
\bibliographystyle{acl_natbib}

\end{document}

%% file: std-macros.tex
\newcommand\sC{\ensuremath{\mathcal{C}}}

\newcommand\sR{\ensuremath{\mathcal{R}}}



%% file: abstract.tex
When answering a question, people often draw upon their rich world knowledge in addition to the particular context. Recent work has focused primarily on answering questions given some relevant document or context, and required very little general background. To investigate question answering with prior knowledge, we present \csqa{}: a challenging new dataset for commonsense question answering. 
To capture common sense beyond associations, we extract from \textsc{ConceptNet} \cite{speer2017conceptnet} multiple target concepts that have the same semantic relation to a single source concept. Crowd-workers are asked to author multiple-choice questions that mention the source concept and discriminate in turn between each of the target concepts. This encourages workers to create questions with complex semantics that often require prior knowledge.
We create 12,247 questions through this procedure and demonstrate the difficulty of our task with a large number of strong baselines. Our best baseline is based on BERT-large \cite{devlin2018BERT} and obtains 56\% accuracy, well below human performance, which is 89\%.

%% file: introduction.tex
\section{Introduction}
\label{introduction}

\begin{figure}[t]
  \includegraphics[width=\columnwidth]{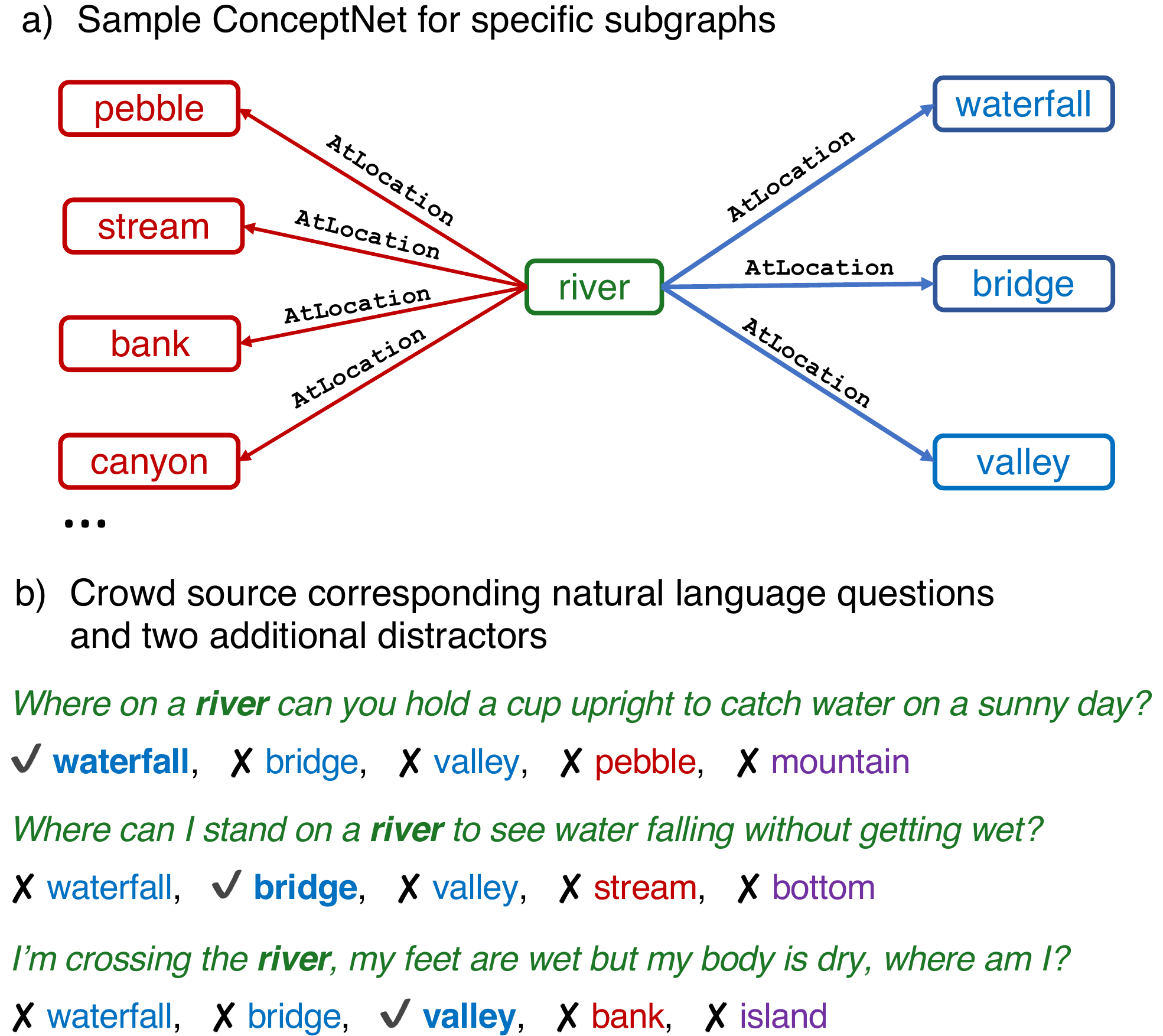}
  \caption{(a) A source concept (in green) and three target concepts (in blue) are sampled from \conceptnet{} (b) Crowd-workers generate three questions, each having one of the target concepts for its answer (\cmark), while the other two targets are not (\xmark). Then, for each question, workers choose an additional distractor from \conceptnet{} (in red), and author one themselves (in purple).
  }~\label{fig:framework}
\end{figure}

When humans answer questions, they capitalize on their common sense and background knowledge about spatial relations, causes and effects, scientific facts and social conventions. 
For instance, given the question \nlex{Where was Simon when he heard the lawn mower?}, one can infer that the lawn mower is close to Simon, and that it is probably outdoors and situated at street level. This type of knowledge seems trivial for humans, but is still out of the reach of current natural language understanding (NLU) systems.

Work on Question Answering (QA) has mostly focused on answering factoid questions, where the answer can be found in a given context with little need for commonsense knowledge \cite{hermann2015teaching,rajpurkar2016squad,nguyen2016ms,joshi2017triviaqa}. 
Small benchmarks such as the Winograd Scheme Challenge \cite{Levesque2011TheWS} and COPA \cite{roemmele2011choice}, targeted common sense more directly, but have been difficult to collect at scale.

Recently, efforts have been invested in developing large-scale datasets for commonsense reasoning. In SWAG \cite{zellers2018swag}, given a textual description of an event, a probable subsequent event needs to be inferred. However, it has been quickly realized that models trained on large amounts of unlabeled data \cite{devlin2018BERT} capture well this type of information and performance on SWAG is already at human level. 
VCR \cite{zellers2018recognition} is another very recent attempt that focuses on the visual aspects of common sense.
Such new attempts highlight the breadth of commonsense phenomena, and make it evident that research on common sense has only scratched the surface. Thus, there is need for datasets and models that will further our understanding of what is captured by current NLU models, and what are the main lacunae.

\ignore{
Work on Question Answering (QA) has mostly focused on factoid QA \cite{hermann2015teaching,rajpurkar2016squad,nguyen2016ms,joshi2017triviaqa}, where an answer is extracted from a textual context using relatively little external knowledge. 
Other small benchmarks, such as the Winograd Scheme Challenge \cite{Levesque2011TheWS} and COPA \cite{roemmele2011choice}, targeted common sense more directly, but have been difficult to collect at scale. Recently, a larger dataset, SWAG, tackled commonsense knowledge about situations \cite{zellers2018swag}. however, as it focuses on sequential grounded situations, it does not capture the full breadth of commonsense types employed by humans.

Moreover, it has become increasingly evident recently \cite{poliak2018hypothesis,gururangan2018annotation}, that dataset generation creates annotation artifacts that are difficult to remove and are specific to the annotation process. In particular, large pre-trained language models, that are fine-tuned on a target task, can obtain surprisingly high performance on datasets such as SWAG and GLUE \cite{devlin2018BERT,wang2018glue}.
This emphasizes the need to create data generation procedures for broad types of common sense that requires complex reasoning, and thus challenges current NLU models. 
}
In this work, we present \csqa{}, a new dataset focusing on commonsense question answering, based on knowledge encoded in \conceptnet{} \cite{speer2017conceptnet}.
We propose a method for generating commonsense questions at scale by asking crowd workers to author questions that describe the relation between concepts from  \conceptnet{} (Figure \ref{fig:framework}). A crowd worker observes a source concept (\emph{`River'} in Figure  \ref{fig:framework}) and three target concepts (\emph{`Waterfall'}, \emph{`Bridge'}, \emph{`Valley'}) that are all related by the same \conceptnet{} relation (\texttt{AtLocation}). The worker then authors three questions, one per target concept, such that only that particular target concept is the answer, while the other two distractor concepts are not. This primes the workers to add commonsense knowledge to the question, that separates the target concept from the distractors. Finally, for each question, the worker chooses one additional distractor from \conceptnet{}, and authors another distractor manually. Thus, in total, five candidate answers accompany each question.

Because questions are generated freely by workers, they often require background knowledge that is trivial to humans but is seldom explicitly reported on the web due to reporting bias \cite{Gordon:2013:RBK:2509558.2509563}. Thus, questions in \csqa{} have a different nature compared to prior QA benchmarks, where
questions are authored given an input text.

Using our method, we collected 12,247 commonsense questions. We present an analysis that illustrates the uniqueness of the gathered questions compared to prior work, and the types of commonsense skills that are required for tackling it.
We extensively evaluate models on \csqa{}, experimenting with pre-trained models, fine-tuned models, and reading comprehension (RC) models that utilize web snippets extracted from Google search on top of the question itself. We find that fine-tuning \textsc{BERT-large} \cite{devlin2018BERT} on \csqa{} obtains the best performance, reaching an accuracy of 55.9\%. This is substantially lower than human performance, which is 88.9\%.

To summarize, our contributions are:
\begin{enumerate}[topsep=0pt,itemsep=0ex,parsep=0ex]
    \item A new QA dataset centered around common sense, containing 12,247 examples.
    \item A new method for generating commonsense questions at scale from \conceptnet{}.
    \item An empirical evaluation of state-of-the-art NLU models on \csqa{}, showing that humans substantially outperform current models.
\end{enumerate}
The dataset can be downloaded from  \url{www.tau-nlp.org/commonsenseqa}. The code for all our baselines 
is available at \url{github.com/jonathanherzig/commonsenseqa}.

%% file: related_work.tex
\section{Related Work}
\label{sec:related_work}

Machine common sense, or the knowledge of and ability to reason about an open ended world, has long been acknowledged as a critical component for natural language understanding. Early work sought programs that could reason about an environment in natural language \cite{mccarthy1959programs}, or leverage a world-model for deeper language understanding \cite{winograd1972language}. Many commonsense representations and inference procedures have been explored \cite{McCHay69,Kowalski:1986:LCE:10030.10034} and large-scale commonsense knowledge-bases have been developed \cite{Lenat1995CYCAL,speer2017conceptnet}. However, evaluating the degree of common sense possessed by a machine remains difficult.

One important benchmark, the Winograd Schema Challenge \cite{Levesque2011TheWS}, asks models to correctly solve paired instances of coreference resolution. 
While the Winograd Schema Challenge remains a tough dataset, the difficulty of generating examples has led to only a small available collection of 150 examples.
The Choice of Plausible Alternatives (COPA) is a similarly important but small dataset consisting of 500 development and 500 test questions \cite{roemmele2011choice}. Each question asks which of two alternatives best reflects a cause or effect relation to the premise. For both datasets, scalability is an issue when evaluating modern modeling approaches.

With the recent adoption of crowdsourcing, several larger datasets have emerged, focusing on predicting relations between situations or events in natural language. JHU Ordinal Commonsense Inference requests a label from 1-5 for the plausibility that one situation entails another \cite{Zhang2017OrdinalCI}. The Story Cloze Test (also referred to as ROC Stories) pits ground-truth endings to stories against implausible false ones \cite{mostafazadeh2016corpus}. Interpolating these approaches, Situations with Adversarial Generations (SWAG), asks models to choose the correct description of what happens next after an initial event \cite{zellers2018swag}. LM-based techniques achieve very high performance on the Story Cloze Test and \textsc{SWAG} by fine-tuning a pre-trained LM on the target task \cite{radford2018improving,devlin2018BERT}. 

Investigations of commonsense datasets, and of natural language datasets more generally, have revealed the difficulty in creating benchmarks that measure the understanding of a program rather than its ability to take advantage of distributional biases, and to model the annotation process \cite{gururangan2018annotation,poliak2018hypothesis}. Annotation artifacts in the Story Cloze Test, for example, allow models to achieve high performance while only looking at the proposed endings and ignoring the stories \cite{Schwartz2017TheEO,Cai2017PayAT}. Thus, the development of benchmarks for common sense remains a difficult challenge.

Researchers have also investigated question answering that utilizes common sense. Science questions often require common sense, and have recently received attention \cite{clark2018think,mihaylov2018can, Ostermann2018MCScriptAN}; however, they also need specialized scientific knowledge. 
In contrast to these efforts, our work studies common sense without requiring additional information. SQUABU
created a small hand-curated test of common sense and science questions \cite{davis2016write}, which are difficult for current techniques to solve. In this work, we create similarly well-crafted questions but at a larger scale.

\nocite{Tange2011a}

%% file: dataset_gen.tex
\section{Dataset Generation}
\label{sec:data_gen}

\begin{figure}[h]
  \center
  \includegraphics[width=1.0\columnwidth]{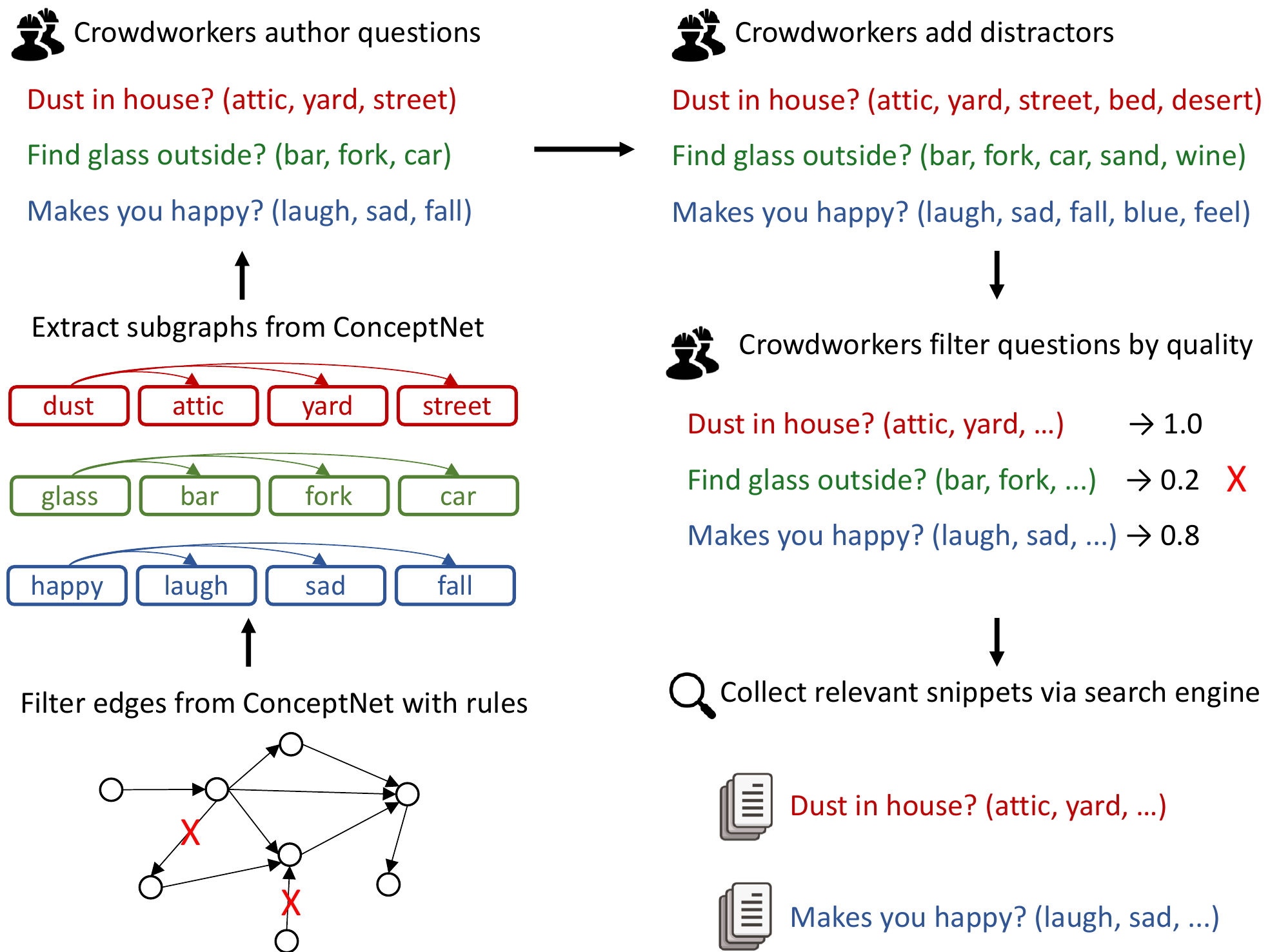}
  \caption{\csqa{} generation process. The input is \conceptnet{} knowledge base, and the output is a set of multiple-choice questions with corresponding relevant context (snippets).}~\label{fig:data_gen}
\end{figure}

Our goal is to develop a method for generating questions that can be easily answered by humans without context, and require commonsense knowledge. We generate multiple-choice questions in a process that comprises the following steps.
\begin{enumerate}[topsep=0pt,itemsep=0ex,parsep=0ex]
    \item We extract subgraphs from \conceptnet{}, each with one source concept and three target concepts.
    \item We ask crowdsourcing workers to author three questions per subgraph (one per target concept), to add two additional distractors per question, and to verify questions' quality.
    \item We add textual context to each question by querying a search engine and retrieving web snippets.
\end{enumerate}
The entire data generation process is summarized in Figure~\ref{fig:data_gen}.
We now elaborate on each of the steps:

\paragraph{Extraction from \conceptnet{}}
\conceptnet{} is a graph knowledge-base $G \subseteq \sC \times \sR \times \sC$, where the nodes $\sC$ represent natural language concepts, and edges $\sR$ represent commonsense relations. Triplets $(c_1, r, c_2)$ carry commonsense knowledge such as `(\emph{gambler}, \texttt{CapableOf}, \emph{lose money})'. \conceptnet{} contains 32 million triplets. To select a subset of triplets for crowdsourcing we take the following steps:
\begin{enumerate}[topsep=0pt,itemsep=0ex,parsep=0ex]
    \item We filter triplets with general relations (e.g., \texttt{RelatedTo}) or relations that are already well-explored in NLP (e.g., \texttt{IsA}). In total we use 22 relations.
    \item We filter triplets where one of the concepts is more than four words or not in English.
    \item We filter triplets where the edit distance between $c_1$ and $c_2$ is too low.
\end{enumerate}
This results in a set of 236,208 triplets $(q, r, a)$, where we call the first concept the \emph{question concept} and the second concept the \emph{answer concept}.

We aim to generate questions that contain the question concept and where the answer is the answer concept. To create multiple-choice questions we need to choose \emph{distractors} for each question. Sampling distractors at random from \conceptnet{} is a bad solution, as such distractors are easy to eliminate using simple surface clues. 

To remedy this, we propose to create \emph{question sets}: for each question concept $q$ and relation $r$ we group three different triplets $\{(q, r, a_1), (q, r, a_2), (q, r, a_3)\}$ (see Figure \ref{fig:framework}). This generates three answer concepts that are semantically similar and have a similar relation to the question concept $q$. This primes crowd workers to formulate questions that require background knowledge about the concepts in order to answer the question.

The above procedure generates approximately 130,000 triplets (43,000 question sets), for which we can potentially generate questions.

\paragraph{Crowdsourcing questions}
We used Amazon Mechanical Turk (AMT) workers to generate and validate commonsense questions.

AMT workers saw, for every question set, the question concept and three answer concepts. They were asked to formulate three questions, where all questions contain the question concept. Each question should have as an answer one of the answer concepts, but not the other two.
To discourage workers from providing simple surface clues for the answer, they were instructed to avoid using words that have a strong relation to the answer concept, for example, not to use the word \emph{`open'} when the answer is \emph{`door'}. 

Formulating questions for our task is non-trivial. Thus, we only accept annotators for which at least 75\% of the questions they formulate pass the verification process described below. 

\paragraph{Adding additional distractors}
To make the task more difficult, we ask crowd-workers to add two additional incorrect answers to each formulated question. One distractor is selected from a set of answer concepts with the same relation to the question concept in \conceptnet{} (Figure~\ref{fig:framework}, in red). The second distractor is formulated manually by the workers themselves (Figure~\ref{fig:framework}, in purple). Workers were encouraged to formulate a distractor that would seem plausible or related to the question but easy for humans to dismiss as incorrect. In total, each formulated question is accompanied with five candidate answers, including one correct answer and four distractors.

\paragraph{Verifying questions quality}
We train a disjoint group of workers to verify the generated questions. Verifiers annotate a question as unanswerable, or choose the right answer. Each question is verified by 2 workers, and only questions verified by at least one worker that answered correctly are used. This processes filters out 15\% of the questions.

\paragraph{Adding textual context}
To examine whether web text is useful for answering commonsense questions, we add textual information to each question in the following way: We issue a web query to Google search for every question and candidate answer, concatenating the answer to the question, e.g., \emph{`What does a parent tell their child to do after they've played with a lot of toys? + ``clean room"'}. We take the first 100 result snippets for each of the five answer candidates, yielding a context of 500 snippets per question. Using this context, we can investigate the performance of reading comprehension (RC) models on \csqa{}.

\paragraph{} Overall, we generated 12,247 final examples, from a total of 16,242 that were formulated. The total cost per question is \$0.33. Table~\ref{tab:formulation_stats} describes the key statistics of \csqa{}.

\begin{table}[]
\begin{center}
\footnotesize{
\begin{tabular}{l|c}
 \textbf{Measurement} & \textbf{Value}\\
 \hline
 \# \conceptnet{} distinct question nodes & 2,254 \\ 
 \# \conceptnet{} distinct answer nodes & 12,094 \\
 \# \conceptnet{} distinct nodes & 12,107 \\ 
 \# \conceptnet{} distinct relation lables & 22 \\ 
 average question length (tokens) & 13.41 \\
 long questions (more than 20 tokens) & 10.3\% \\ 
 average answer length (tokens) & 1.5 \\
 \# answers with more than 1 token & 44\% \\
 \# of distinct words in questions & 14,754 \\
 \# of distinct words in answers & 4,911 \\
\end{tabular}}
\end{center}
\caption{Key statistics for \dataset}
\label{tab:formulation_stats}
\end{table}

%% file: dataset_analysis.tex
\section{Dataset Analysis}
\label{sec:data_analysis}

\begin{table*}[t]
\centering
\resizebox{1.0\textwidth}{!}{
\begin{tabular}{l|l|c}
 \textbf{Relation} & \textbf{Formulated question example} & {\%}\\ 
\hline

\texttt{AtLocation} & \emph{Where would I not want a fox?} \textbf{A.} hen house, \textbf{B.} england, \textbf{C.} mountains, \textbf{D.} ... & 47.3  \\ 
\texttt{Causes} &\emph{What is the hopeful result of going to see a play?} \textbf{A.} being entertained, \textbf{B.} meet, \textbf{C.} sit, \textbf{D.} ... & 17.3 \\ 
\texttt{CapableOf} & \emph{Why would a person put flowers in a room with dirty gym socks?} \textbf{A.} smell good, \textbf{B.} many colors, \textbf{C.} continue to grow , \textbf{D.} ... & 9.4 \\ 
\texttt{Antonym} & \emph{Someone who had a very bad flight might be given a trip in this to make up for it?} \textbf{A.} first class, \textbf{B.} reputable, \textbf{C.} propitious , \textbf{D.} ... & 8.5 \\
\texttt{HasSubevent} & \emph{How does a person begin to attract another person for reproducing?} \textbf{A.} kiss, \textbf{B.} genetic mutation, \textbf{C.} have sex , \textbf{D.} ... & 3.6  \\ 
\texttt{HasPrerequisite} & \emph{If I am tilting a drink toward my face, what should I do before the liquid spills over?} \textbf{A.} open mouth, \textbf{B.} eat first, \textbf{C.} use glass , \textbf{D.} ... & 3.3  \\ 
\texttt{CausesDesire} & \emph{What do parents encourage kids to do when they experience boredom?} \textbf{A.} read book, \textbf{B.} sleep, \textbf{C.} travel , \textbf{D.} ... & 2.1  \\
\texttt{Desires} & \emph{What do all humans want to experience in their own home?} \textbf{A.} feel comfortable, \textbf{B.} work hard, \textbf{C.} fall in love , \textbf{D.} ... & 1.7  \\
\texttt{PartOf} & \emph{What would someone wear to protect themselves from a cannon?} \textbf{A.} body armor, \textbf{B.} tank, \textbf{C.} hat , \textbf{D.} ... & 1.6  \\
\texttt{HasProperty} & \emph{What is a reason to pay your television bill?} \textbf{A.} legal, \textbf{B.} obsolete, \textbf{C.} entertaining , \textbf{D.} ... & 1.2  \\
\end{tabular}}
\caption{Top \conceptnet{} relations in \csqa{}, along with their frequency in the data and an example question. The first answer (\textbf{A}) is the correct answer}
\label{tab:cn_relations}
\end{table*}

\paragraph{\conceptnet{} concepts and relations}

\csqa{} builds on \conceptnet{}, which contains \emph{concepts} such as \texttt{dog}, \texttt{house}, or \texttt{row boat}, connected by \emph{relations} such as \texttt{Causes}, \texttt{CapableOf}, or \texttt{Antonym}. The top-5 question concepts in \csqa{} are \emph{`Person'} (3.1\%), \emph{`People'} (2.0\%), \emph{`Human'} (0.7\%), \emph{`Water'} (0.5\%) and \emph{`Cat'} (0.5\%). In addition,
we present the main relations along with the percentage of questions generated from them in Table \ref{tab:cn_relations}. It's worth noting that since question formulators were not shown the \conceptnet{} relation, they often asked questions that probe other relationships between the concepts. For example, the question \nlex{What do \textbf{audiences} clap for?} was generated from the \texttt{AtLocation} relation, but focuses on social conventions instead.

\begin{figure}[t]
  \includegraphics[width=1.05\columnwidth]{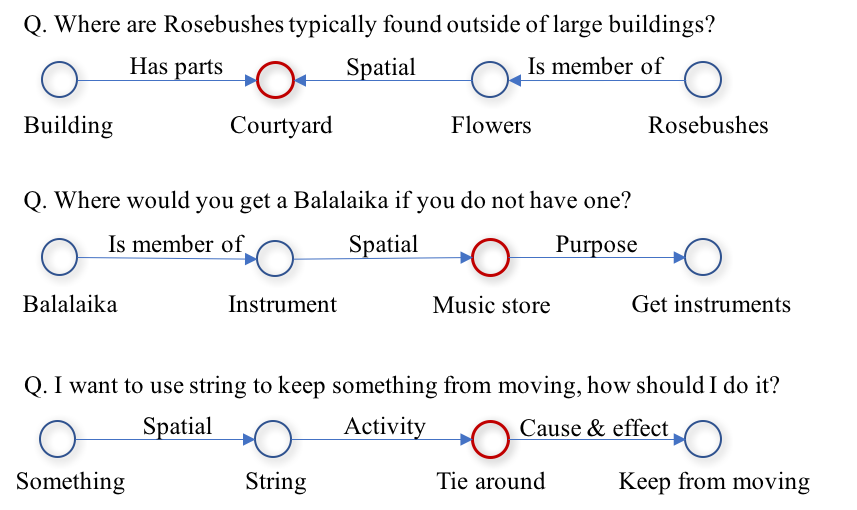}
  \caption{Examples of manually-annotated questions, with the required skills needed to arrive at the answers (red circles). Skills are labeled edges, and concepts are nodes.
  }~\label{fig:skills_examples}
\end{figure}

\paragraph{Question formulation}
Question formulators were instructed to create questions with high language variation. 122 formulators contributed to question generation. However, 10 workers formulated more than 85\% of the questions.

\begin{figure*}[h]
    \center
  \includegraphics[width=0.9\textwidth]{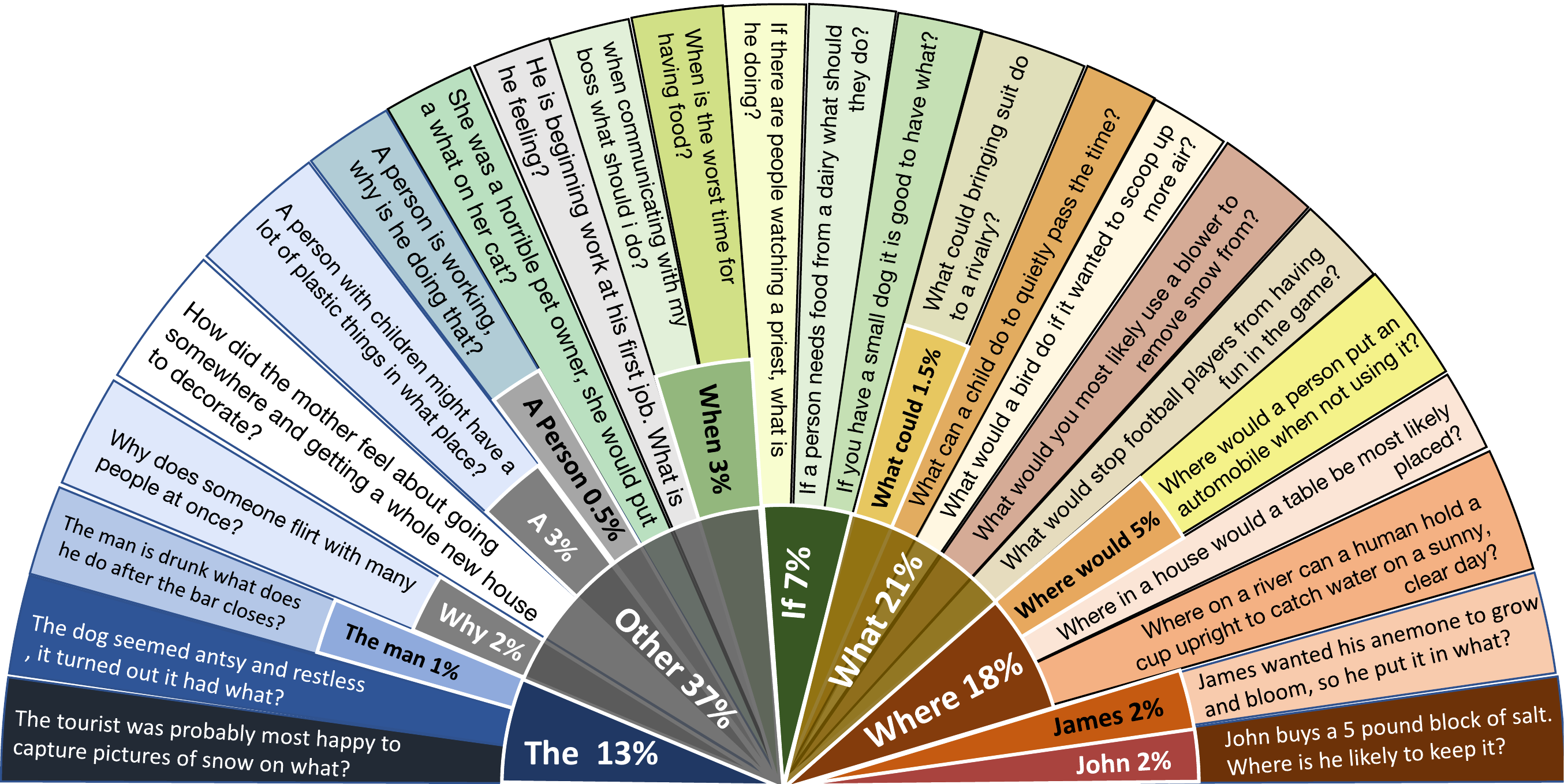}
  \caption{Distribution of the first and second words in questions. The inner part displays words and their frequency and the outer part provides example questions. 
  }~\label{fig:first-word-distrubition}
\end{figure*}

We analyzed the distribution of first and second words in the formulated questions along with example questions. Figure~\ref{fig:first-word-distrubition} presents the breakdown. Interestingly, only 44\% of the first words are WH-words. In about 5\% of the questions, formulators used first names to create a context story, and in 7\% they used the word \nlex{if} to present a hypothetical question. This suggests high variability in the question language.  

\paragraph{Commonsense Skills}
To analyze the types of commonsense knowledge needed to correctly answer questions in \csqa{}, we randomly sampled 100 examples from the development set and performed the following analysis.

For each question, we explicitly annotated the types of commonsense skills that a human uses to answer the question. We allow multiple commonsense skills per questions, with an average of 1.75 skills per question. Figure~\ref{fig:skills_examples} provides three example annotations. Each annotation contains a node for the answer concept, and other nodes for concepts that appear in the question or latent concepts. Labeled edges describe the commonsense skill that relates the two nodes. We defined commonsense skills based on the analysis of \newcite{lobue2011types}, with slight modifications to accommodate the phenomena in our data. Table~\ref{tab:skills} presents the skill categories we used, their definition and their frequency in the analyzed examples.

\begin{table}[t]
\centering
\resizebox{1.0\columnwidth}{!}{
\begin{tabular}{l|l|c}
Category         & Definition                                                   & \% \\
\hline
Spatial          & Concept A appears near Concept B                             & 41      \\
Cause \& Effect  & Concept A causes Concept B                                     & 23      \\
Has parts        & Concept A contains Concept B as one of its parts                & 23      \\
Is member of     & Concept A belongs to the larger class of Concept B             & 17      \\
Purpose          & Concept A is the purpose of Concept B                          & 18      \\
Social           & It is a social convention that Concept A                 & 15      \\
                 & correlates with Concept B  & \\
Activity         & Concept A is an activity performed in the context            & 8 \\
                 & of Concept B  & \\
Definition       & Concept A is a definition of Concept B                         & 6 \\
Preconditions    & Concept A must hold true in order for Concept B to             & 3 \\
                 & take place & \\
\end{tabular}}
\caption{Skills and their frequency in the sampled data. As each example can be annotated with multiple skills, the total frequency does not sum to 100\%.  }
\label{tab:skills}
\end{table}

%% file: baselines.tex
\section{Baseline Models}
\label{sec:base}


Our goal is to collect a dataset of commonsense questions that are easy for humans, but hard for current NLU models. 
To evaluate this, we experiment with multiple baselines.
Table \ref{tab:models} summarizes the various baseline types and characterizes them based on (a) whether training is done on \csqa{} or the model is fully pre-trained, and (b) whether context (web snippets) is used. We now elaborate on the different baselines.





\begin{table}[t]
\centering
\resizebox{0.5\columnwidth}{!}{
\begin{tabular}{lccc}
\hline \hline
Model        & Training & Context \\ \hline
\textsc{VecSim}          & \xmark          & \xmark           \\
\textsc{LM1B}           & \xmark          & \xmark           \\
\textsc{QABilinear}                 & \cmark        & \xmark         \\
\textsc{QACompare}                 & \cmark        & \xmark         \\
\textsc{ESIM}                 & \cmark        & \xmark         \\
\textsc{GPT}                 & \cmark        & \xmark         \\
\textsc{BERT}                 & \cmark        & \xmark         \\
\textsc{BIDAF++}                & \cmark        & \cmark         \\
\hline \hline      
\end{tabular}}
\caption{Baseline models along with their characteristics. \textit{Training} states whether the model was trained on \csqa{}, or was only trained a different dataset. \textit{Context} states whether the model uses extra context as input.}
\label{tab:models}
\end{table}

\begin{enumerate}[wide, labelwidth=!,labelindent=0pt,topsep=0pt,itemsep=0pt,label=\textbf{\alph*}]
    \item \textbf{\textsc{VecSim}} A  model that chooses the answer with highest cosine similarity to the question, where the question and answers are represented by an average of pre-trained word embeddings.
    \item \textbf{\textsc{LM1B}} Inspired by \newcite{trinh2018simple}, we employ a large language model (LM) from \newcite{jozefowicz2016exploring}, which was pre-trained on the One Billion Words Benchmark \cite{chelba2013one}. We use this model in two variations. In the first (\textsc{LM1B-concat}), we simply concatenate each answer to the question. In the second (\textsc{LM1B-rep}), we first cluster questions according to their first two words. Then, we recognize five high-frequency prefixes that cover 35\% of the development set (e.g., \nlex{what is}). We rephrase questions that fit into one of these prefixes as a declarative sentence that contains the answer. E.g., we rephrase \nlex{What is usually next to a door?} and the candidate answer \nlex{wall} to \nlex{Wall is usually next to a door}.
    For questions that do not start with the above prefixes, we concatenate the answer as in \textsc{LM1B-concat}. In both variations we return the answer with highest LM probability.
    \item \textbf{\textsc{QABilinear}} This model, propsed by \newcite{yu2014deep} for QA, scores an answer $a_i$ with a bilinear model: $q W a_i^\top$, where the question $q$ and answers $a_i$ are the average pre-trained word embeddings and $W$ is a learned parameter matrix. A softmax layer over the candidate answers is used to train the model with cross-entropy loss.
    \item \textbf{\textsc{QACompare}} This model is similar to an NLI model from \newcite{liu2016learning}. The model represents the interaction between the question $q$ and a candidate answer $a_i$ as:  $h=\text{relu}([q ; a_i ; q \odot a_i; q - a_i]W_1 + b_1)$, where '$;$' denotes concatenation and $\odot$ is element-wise product. Then, the model predicts an answer score using a feed forward layer: $hW_2 + b_2$. Average pre-trained embeddings and softmax are used to train the model.
    
    \item \textbf{\textsc{ESIM}} We use ESIM, a strong NLI model \cite{chen2016enhanced}. Similar to \newcite{zellers2018swag}, we change the output layer size to the number of candidate answers, and apply softmax to train with cross-entropy loss.


    \item \textbf{\textsc{BiDAF++}} A state-of-the-art RC model, that uses the retrieved Google web snippets (Section~\ref{sec:data_gen}) as context. We augment \textsc{BiDAF} \cite{seo2016bidaf} with a self-attention layer and ELMo representations \cite{Peters:2018,huang2018flowqa}.
    To adapt to the multiple-choice setting, we choose the answer with highest model probability.
    
    \item \textbf{\textsc{Generative Pre-trained Transformer (GPT)}} \newcite{radford2018improving} proposed a method for adapting pre-trained LMs to perform a wide range of tasks. We applied their model to \dataset{} by encoding each question and its candidate answers as a series of delimiter-separated sequences. For example, the question \nlex{If you needed a lamp to do your work, where would you put it?}, and the candidate answer \nlex{bedroom} would become ``\texttt{[start]} \emph{If ... ?} \texttt{[sep]} \emph{bedroom} \texttt{[end]}". The hidden representations over each \texttt{[end]} token are converted to logits by a linear transformation and passed through a softmax to produce final probabilities for the answers. We used the same pre-trained LM and hyper-parameters for fine-tuning as \newcite{radford2018improving} on ROC Stories, except with a batch size of 10.

    \item \textbf{\textsc{BERT}} Similarly to the \textsc{GPT}, BERT fine-tunes a language model and currently holds state-of-the-art across a broad range of tasks \cite{devlin2018BERT}. \textsc{BERT} uses a masked language modeling objective, which predicts missing words masked from unlabeled text. 
    To apply \textsc{BERT} to \dataset{}, we linearize each question-answer pair into a delimiter-separated sequence (i.e., ``\texttt{[CLS]} \emph{If ... ?} \texttt{[SEP]} \emph{bedroom} \texttt{[SEP]}") then fine-tune the pre-trained weights from uncased \textsc{BERT-large}.\footnote{The original weights and code released by Google may be found here: https://github.com/google-research/bert}
    Similarly to the GPT, the hidden representations over each \texttt{[CLS]} token are run through a softmax layer to create the predictions.
    We used 
    the same hyper-parameters as \newcite{devlin2018BERT} for SWAG.
    
\end{enumerate}

%% file: experiments.tex
\section{Experiments}

\paragraph{Experimental Setup}
We split the data into a training/development/test set with an 80/10/10 split. We perform two types of splits: (a) \emph{random split} --  where questions are split uniformly at random, and (b) \emph{question concept split} -- where each of the three sets have disjoint question concepts. We empirically find (see below) that a random split is harder for models that learn from \csqa{}, because the same question concept appears in the training set and development/test set with different answer concepts, and networks that memorize might fail in such a scenario. Since the random split is harder, we consider it the primary split of \csqa{}.

We evaluate all models on the test set using accuracy (proportion of examples for which prediction is correct), and tune hyper-parameters for all trained models on the development set.
To understand the difficulty of the task, we add a SANITY mode, where we replace the hard distractors (that share a relation with the question concept and one formulated by a worker) with random \conceptnet{} distractors. We expect a reasonable baseline to perform much better in this mode.

For pre-trained word embeddings we consider
300d GloVe embeddings \cite{pennington2014glove} and 300d Numberbatch \conceptnet{} node embeddings \cite{speer2017conceptnet}, which are kept fixed at training time.
We also combine \textsc{ESIM} with 1024d ELMo contextual representations, which are also fixed during training.

\begin{table*}[t]
\centering
\resizebox{0.6\textwidth}{!}{
\begin{tabular}{l|cc|cc}
                                                 & \multicolumn{2}{c|}{Random split} & \multicolumn{2}{c}{Question concept split} \\ \cline{1-5}
Model                                            & Accuracy         & SANITY         & Accuracy        & SANITY         \\ \cline{1-5}
\textsc{VecSim+Numberbatch}     & 29.1           & 54.0         & 30.3            & 54.9        \\
\textsc{LM1B-rep}               & 26.1           & 39.6            & 26.0            & 39.1            \\
\textsc{LM1B-concat}            & 25.3           & 37.4            & 25.3            & 35.2    \\
\textsc{VecSim+GloVe}           &22.3           & 26.8         & 20.8           & 27.1        \\
\hline
\textsc{BERT-large}                    & \textbf{55.9}  & \textbf{92.3}         & \textbf{63.6}   & \textbf{93.2}            \\
\textsc{GPT}                    & 45.5  & 87.2            & 55.5   & 88.9            \\
\textsc{ESIM+ELMo}              & 34.1  & 76.9            & 37.9     & 77.8            \\
\textsc{ESIM+GloVe}             & 32.8  & 79.1            & 40.4   & 78.2            \\
\textsc{QABilinear+GloVe}       & 31.5           & 74.8         & 34.2            & 71.8        \\
\textsc{ESIM+Numberbatch}       & 30.1           & 74.6            & 31.2            & 75.1            \\
\textsc{QABilinear+Numberbatch} & 28.8           & 73.3         & 32.0            & 71.6        \\
\textsc{QACompare+GloVe}        & 25.7           & 69.2         & 34.1            & 71.3        \\
\textsc{QACompare+Numberbatch}  & 20.4           & 60.6         & 25.2            & 66.8        \\
\hline
\textsc{BiDAF++}               & 32.0           & 71.0            & 38.4            & 72.0            \\
\hline \hline
\textsc{Human}                  & \textbf{88.9}  &            &                  &              \\ \cline{1-5}
\end{tabular}}
\caption{Test set accuracy for all models.}
\label{tab:main_res}
\end{table*}

\begin{table*}[t]
\centering
\resizebox{1.0\textwidth}{!}{
\begin{tabular}{l|l|l|l|l|l}
 \textbf{Category} & \textbf{Formulated question example} & \textbf{Correct answer} & \textbf{Distractor} & \textbf{Accuracy} & \textbf{\%} \\ 
\hline

Surface & \emph{If someone laughs after surprising them they have a good sense of what?} & \textbf{humor} & laughter & 77.7 & 35\% \\ 
clues &\emph{How might a automobile get off a freeway?} & \textbf{exit ramp} & driveway  & &  \\ 
\hline
Negation / & \emph{Where would you store a pillow case that is not in use?} & drawer & \textbf{bedroom} & 42.8 & 7\%  \\
Antonym & \emph{Where might the stapler be if I cannot find it?}  & desk drawer & \textbf{desktop} &  &  \\
\hline
 Factoid & \emph{How many hours are in a day?} & twenty four & \textbf{week} & 38.4 & 13\%  \\
knowledge & \emph{What geographic area is a lizard likely to be?}  & west texas & \textbf{ball stopped} &  &  \\
\hline
Bad & \emph{Where is a well used toy car likely to be found?} & child's room & \textbf{own home} & 35.4 & 31\%  \\
granularity & \emph{Where may you be if you're buying pork chops at a corner shop?}  & iowa & \textbf{town} &  &  \\
\hline
Conjunction & \emph{What can you use to store a book while traveling?}  & suitcase & \textbf{library of congress} & 23.8 & 23\% \\
 & \emph{On a hot day what can you do to enjoy something cool and sweet?} & eat ice cream & \textbf{fresh cake} &  & \\

\end{tabular}}
\caption{\textsc{BERT-large} baseline analysis. For each category we provide two examples, the correct answer, one distractor, model accuracy and frequency in the dataset. The predicted answer is in bold.}
\label{tab:model_analysis}
\end{table*}

\paragraph{Human Evaluation}
To test human accuracy, we created a separate task for which we did not use a qualification test, nor used AMT master workers. We sampled 100 random questions and for each question gathered answers from five workers that were not involved in question generation. 
Humans obtain 88.9\% accuracy, taking a majority vote for each question.
\ignore{
and taking all answers we obtain 82\% human accuracy (since workers in this task are untrained, we use a majority vote to mitigate the effect of malicious workers). }

\paragraph{Results}
Table~\ref{tab:main_res} presents test set results for all models and setups.

The best baselines are \textsc{BERT-large} and \textsc{GPT} with an accuracy of 55.9\% and 45.5\%, respectively, on the random split (63.6\% and 55.5\%, respectively, on the question concept split). This is well below human accuracy, demonstrating that the benchmark is much easier for humans. Nevertheless, this result is much higher than random (20\%), showing the ability of language models to store large amounts of information related to commonsense knowledge.

The top part of Table~\ref{tab:main_res} describes untrained models. We observe that performance is higher than random, but still quite low. The middle part describes models that were trained on \csqa{}, where \textsc{BERT-large} obtains best performance, as mentioned above. 
\textsc{ESIM} models follow \textsc{BERT-large} and \textsc{GPT}, and obtain much lower performance. We note that ELMo representations did not improve performance compared to GloVe embeddings, possibly because we were unable to improve performance by back-propagating into the representations themselves (as we do in \textsc{BERT-large} and \textsc{GPT}). 
The bottom part shows results for \textsc{BiDAF++} that uses web snippets as context. We observe that using snippets does not lead to high performance, hinting that they do not carry a lot of useful information.

Performance on the random split is five points lower than the question concept split on average across all trained models. We hypothesize that this is because having questions in the development/test set that share a question concept with the training set, but have a different answer, creates difficulty for networks that memorize the relation between a question concept and an answer.

Lastly, all SANITY models that were trained on \csqa{} achieve very high performance (92\% for \textsc{BERT-large}), showing that selecting difficult distractors is crucial.

\paragraph{Baseline analysis}

To understand the performance of \textsc{BERT-large}, we analyzed 100 examples from the development set (Table~\ref{tab:model_analysis}). We labeled examples with categories (possibly more than one per example) and then computed the average accuracy of the model for each category.

We found that the model does well (77.7\% accuracy) on examples where surface clues hint to the correct answer. Examples that involve negation or understanding antonyms have lower accuracy (42.8\%), similarly to examples that require factoid knowledge (38.4\%). Accuracy is particularly low in questions where the correct answer has finer granularity compared to one of the distractors (35.4\%), and in cases where the correct answer needs to meet a conjunction of conditions, and the distractor meets only one of them (23.8\%).

\begin{figure}[t]
  \includegraphics[width=\columnwidth,trim={0 0 0 1.2cm},clip]{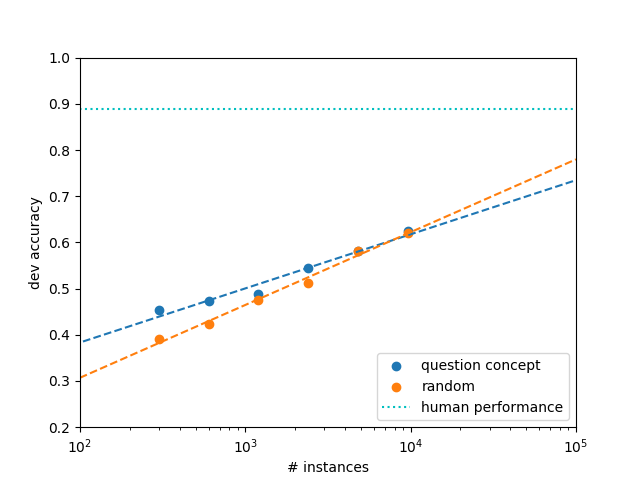}
  \caption{Development accuracy for \textsc{BERT-large} trained with varying amounts of data.}~\label{fig:learning_curves}
\end{figure}

\paragraph{Learning Curves}
To extrapolate how current models might perform with more data, we evaluated \text{BERT-large} on the development set, training with varying amounts of data. The resulting learning curves are plotted in figure \ref{fig:learning_curves}. For each training set size, hyper-parameters were identical to section \ref{sec:base}, except the number of epochs was varied to keep the number of mini-batches during training constant. To deal with learning instabilities, each data point is the best of 3 runs. We observe that the accuracy of \textsc{BERT-large} is expected to be roughly 75\% assuming 100k examples, still substantially lower than human performance.

%% file: conclusion.tex
\section{Conclusion}

We present \csqa{}, a new QA dataset that contains 12,247 examples and aims to test commonsense knowledge. We describe a process for generating difficult questions at scale using \conceptnet{}, perform a detailed analysis of the dataset, which elucidates the unique properties of our dataset, and extensively evaluate on a strong suite of baselines. We find that the best model is a pre-trained LM tuned for our task and obtains 55.9\% accuracy, dozens of points lower than human accuracy. We hope that this dataset facilitates future work in incorporating commonsense knowledge into NLU systems.